\documentclass[journal,onecolumn]{IEEEtran}
\ifCLASSINFOpdf
\else
\fi

\usepackage{epsfig} 
   \usepackage{tikz}
   \usepackage{tikz-3dplot}
   \usepackage{amsmath} 
   \usepackage{amssymb,tabularx,siunitx}  
   
\newcommand{\mr}[1]{\mathrm{#1}}
\newcommand{\bs}[1]{\boldsymbol{#1}}

\usepackage{bm}
\newcommand{\beq}{\begin{equation}}
\newcommand{\eeq}{\end{equation}}
\newcommand{\btable}{\begin{table}}
\newcommand{\etable}{\end{table}}
\newcommand{\bal}{\begin{aligned}}
\newcommand{\eal}{\end{aligned}}
\newcommand{\beqstar}{\begin{equation*}}
\newcommand{\eeqstar}{\end{equation*}}
\newcommand{\bb}{\begin{bmatrix}}
\newcommand{\eb}{\end{bmatrix}}

\newcommand{\bp}{\begin{pmatrix}}
\newcommand{\ep}{\end{pmatrix}}
\newcommand{\bsa}{\left\{\begin{array}{c}}
\newcommand{\esa}{\end{array}\right\}}
\newcommand{\ba}{\left[\begin{array}}
\newcommand{\ea}{\end{array}\right]}
\newcommand{\bad}{\left|\begin{array}}
\newcommand{\ead}{\end{array}\right|}

\newcommand{\tr}{{\mathrm T}}
\newcommand{\inv}{{-1}}

\newcommand{\ssymbol}[1]{^{\@fnsymbol{#1}}}

\usepackage{algorithm}
\usepackage{algpseudocode}

\begin{document}
%
\title{Vision-based control of a knuckle boom crane with online cable length estimation}
%
%
%

\author{Geir Ole Tysse,
        Andrej~Cibicik
        and~Olav~Egeland 
\thanks{The research presented in this paper has received funding from the Norwegian Research Council, SFI Offshore Mechatronics, project number 237896.}
\thanks{The authors are with the Dept. of Mechanical and Industrial Engineering, Norwegian University of Science and Technology, NO-7465 Trondheim, Norway.
  (e-mail: geir.o.tysse@ntnu.no, andrej.cibicik@ntnu.no and olav.egeland@ntnu.no).}}

%
%

\markboth{}%
{Tysse \MakeLowercase{\textit{et al.}}: Vision-based control of a knuckle boom crane with on-line cable length estimation}
%




\maketitle

\begin{abstract}
A vision-based controller for a knuckle boom crane is presented. The controller is used to control the motion of the crane tip and at the same time compensate for payload oscillations. The oscillations of the payload are measured with three cameras that are fixed to the crane king and are used to track two spherical markers fixed to the payload cable. Based on color and size information, each camera identifies the image points corresponding to the markers. The payload angles are then determined using linear triangulation of the image points. An extended Kalman filter is used for estimation of payload angles and angular velocity. The length of the payload cable is also estimated using a least squares technique with projection. The crane is controlled by a linear cascade controller where the inner control loop is designed to damp out the pendulum oscillation, and the crane tip is controlled by the outer loop. The control variable of the controller is the commanded crane tip acceleration, which is converted to a velocity command using a velocity loop. 
The performance of the control system is studied experimentally using a scaled laboratory version of a knuckle boom crane. 
\end{abstract}


%
\IEEEpeerreviewmaketitle

\section{Introduction}
\label{intro}
\IEEEPARstart{C}{ranes} are important in a wide range of operations both onshore and offshore. Crane hoisting operations may be associated with high risk due to the motion of a heavy payload. The landing of the payload is especially critical, since underestimation of the payload motion may lead to damage of equipment and injuries to personnel on the landing site. In addition, the knowledge of the vertical position of the payload in relation to the landing site is necessary. Currently, most of the cranes are driven manually by an operator, and without automation for suppressing the sway of the payload. Automatic control of cranes may contribute to safety of crane operations and reduction of delays. In this work we present a mechatronic system including a method for the estimation of the crane payload motion using a vision-based system, a controller for suppressing the payload sway, and a procedure for cable length estimation using an adaptive law. 

Crane control has been an active area of research during the last decades \cite{Abdel-Rahman2003,Ramli2017}. One approach is open loop control laws. This includes input shaping, which was studied in \cite{Blackburn2010}, and feedforward techniques \cite{Kim2010}. Such methods do not require a feedback signal and can be effective, however they require that the model of the system is sufficiently accurate. Another approach is closed loop methods that are less sensitive to modeling errors and noise.  
A number of energy-based controllers for damping the payload oscillations have been proposed for a 2-DOF cart-payload system. In \cite{Yu1995}  a controller based on singular perturbation techniques  was proposed, while in \cite{Yoshida1998} a Lyapunov-based controller with constrained rope length and trolley stroke was proposed. In \cite{Vyhlidal2017} the authors proposed a time-delay feedback controller for suppressing the payload oscillation by adjusting the cable length. In \cite{Sun2015} the authors proposed an energy-based controller for trolley position and hoisting in combination with damping of the payload oscillation in one plane. Controllers have also been proposed for overhead cranes modeled with more degrees of freedom. In \cite{Sun2013} an energy coupling based output feedback controller for a 4-DOF overhead crane was presented. In \cite{Cibicik2018} a controller based on LaSalle's invariance principle for a bifilar payload and cart system was presented. Controllers based on feedback linearization for damping payload oscillation for 3-DOF overhead cranes were proposed in \cite{Chwa2009}.  Model predictive controller (MPC) is another type of a closed loop controller that has been used in crane control by several authors. MPC was used for controlling a 1-DOF trolley with a pendulum in \cite{Wu2015} and \cite{Vukov2012}. In \cite{Arnold2005} a real-time MPC was proposed for a linearized model of a mobile harbour crane, where the luffing and slewing motion were controlled. This work was further developed in \cite{Neupert2010} with a stabilizing feedback part, linearized feedforward part, and where flatness was used for simplifying the model. 
The reader can refer to the review papers \cite{Abdel-Rahman2003} and \cite{Ramli2017}, where crane control strategies are extensively discussed. 

A knuckle boom crane has a kinematic structure which is similar to a robotic manipulator, which means that results from robot control can be adapted to the control of knuckle boom cranes. One example of this is 2D cameras, which are widely used in robotics. Vision allows a robotic system to obtain geometrical information from the surrounding environment to be used for motion planning and control \cite{zhang2018unified}. In visual servoing the end-effector is controlled relative to a target using visual features extracted from one or several cameras. Early work in this field is well covered in \cite{Hutchinson1996}. There are mainly two different visual servoing approaches: image-based visual servoing and position-based visual servoing. In image-based visual servoing \cite{Espiau1992} the error is defined in the 2D image space, and in position-based visual servoing \cite{Thuilot2002} the error is defined in the 3D Cartesian space. Comparison of these methods are found e.g., in \cite{Janabi2011}. A challenge in position-based visual servoing is that the information about metric distances is lost in the camera projection \cite{Chiuso2002}. Stereo vision is one of the approaches that can be used for recovering of metric distances \cite{Panahandeh2014}. In \cite{Weng1993} the authors used epipolar geometry for two cameras with a nonlinear minimization technique for recovering the metric distances. In \cite{Pereira2018} a two-view bundle adjustment approach for visual navigation was presented. The 3D scene points can be extracted from the scene objects, which are viewed as image correspondences. A correspondence is a pair of corresponding features in different images that represent the same scene object. A method for determining 3D position of points from correspondences is called triangulation, which was solved for the two-view problem in \cite{Hartley2004}. Review of stereo vision for tracking can be found in \cite{Scharstein2002}, \cite{Brown2003} and \cite{Tippetts2016}. In the case of a three-view configuration the accuracy of estimates can be improved, but also the complexity is increased \cite{Stewenius2005}. For a three-view case the triangulation was solved by optimisation using epipolar geometry constraints in \cite{Kukelova2013} and by a proposed iterative nonlinear least-square solver in \cite{Hedborg2014}.

Adaptive parameter estimation and system identification have been studied for models with unknown parameters in \cite{Sastry1989} and \cite{ioannou2012}. 
Algorithms based on projection, least-squares and gradient search have been widely used, where unknown parameters were estimated using adaptive laws. Online parameter estimation have been used for several types of systems. In \cite{Kenne2008} and \cite{Na2015} the authors discussed different adaptive estimation of time-varying parameter approaches in nonlinear systems. 
In \cite{Lertpiriyasuwat2006} the authors proposed a method for adaptive real-time estimation of the pose of the end-effector of an industrial robot. In \cite{Shin2018} the authors proposed using an adaptive range estimation for a vision system on a UAV.   

A typical assumption in crane control papers is that the cable length is known. In practice, it may be the case that only the cable length from the crane tip to the hook is known, while the payload is suspended down from the hook with slings or chains of unknown length. The information about the total cable length is required both for the controller and the estimator of payload states, therefore estimation of the cable length is an important task. Although knuckle boom cranes are widely used for marine vessels, there is little experimental results on automatic control of this type of cranes in the research literature.

%
\begin{figure}
\centering
  \begin{tikzpicture}[scale=1.0]
      \node at (0,0) {\includegraphics[width=0.35\textwidth]{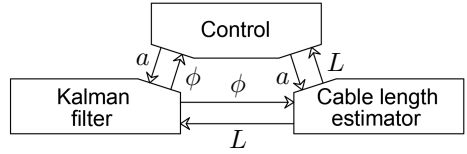}};
  \node at (37pt,3pt) {$L$};
  \node at (0pt,-6pt) {$\phi$};
  \node at (-17pt,-3pt) {$\phi$};
  \node at (17pt,-3pt) {$a$};
  \node at (0pt,-26pt) {$L$};
  \node at (-36pt,4pt) {$a$};
  \end{tikzpicture}
\caption{Communication between the controller, the extended Kalman filter and the cable length estimator. The signals are denoted as follows: $a$ are the commanded control accelerations, $\phi$ are the estimates of the payload states and $L$ is the estimate of the cable length.}
\label{depend_fig}
\end{figure}
\begin{figure}
\centering
\includegraphics[width=0.47\textwidth]{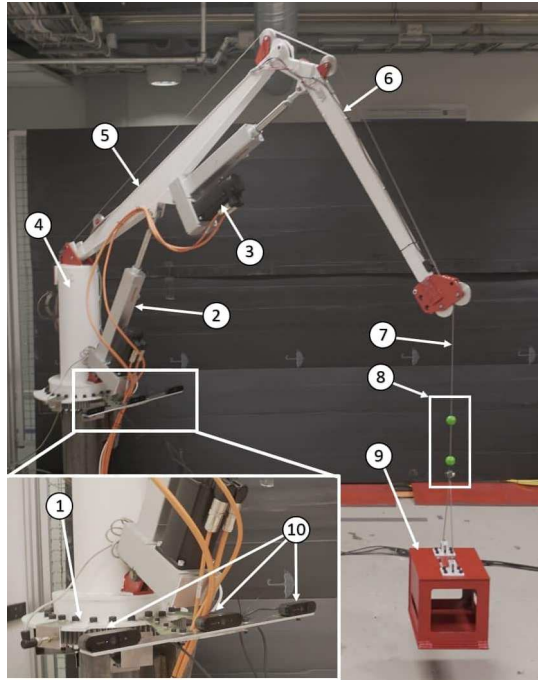}
\caption{A small-scale knuckle boom crane with a vision-based sensor: (1) slew joint $q_1$, (2) linear actuator $q_2$, (3) linear actuator $q_3$, (4) crane king, (5) inner boom, (6) outer boom, (7) cable, (8) spherical markers, (9) payload and (10) 2D cameras.}
\label{fig:crane_system}
\end{figure}
In this work we present a vision-based control of a knuckle boom crane with online cable length estimation. The measurements of the payload cable orientation angles are obtained by a sensor configuration where three cameras are rigidly attached to the crane king,  which is a new sensor arrangement for crane control. The cameras are used to track the position of two spherical markers attached to the payload cable, where the color and size information of the markers is used. We implement the direct linear transformation procedure, which was  given for stereo vision in \cite{Hartley2004}, for a three-camera case to increase the accuracy of the measurements. An extended Kalman filter is used to estimate the orientation angles and angular velocity of the payload pendulation. Estimation of the cable length is done with a least squares technique with projection based on an adaptive law. The suggested linear cascade controller damps out the payload oscillations and controls the position of the crane tip. The direct control output is the commanded acceleration of the crane tip, which is converted to the velocity signal, which is a common type of control input for cranes, by the velocity loop \cite{Sawodny2002}.  
In the proposed mechatronic system the controller, the payload state estimator and the cable length estimator are interconnected as shown in Fig.~\ref{depend_fig}. 
The performance of the proposed sensor, controller and cable length estimator is studied experimentally using a scaled laboratory version of a knuckle boom crane and realistic geometry of the payload.

The rest of this work is organized as follows. In Section \ref{system} the knuckle boom crane system is presented, as well as necessary kinematic and dynamic derivations are given. In Section \ref{vision} the vision-based payload motion estimator is discussed, while in Section \ref{cable_length} the procedure for cable length estimation is presented. The controller for the crane is derived in Section \ref{control}, the experimental results are given in Section \ref{experiment} and the conclusions are given in Section \ref{conclusions}.

\section{Knuckle boom crane with payload}
\label{system}
\subsection{System description}
\begin{figure}
\centering
\includegraphics[width=0.47\textwidth]{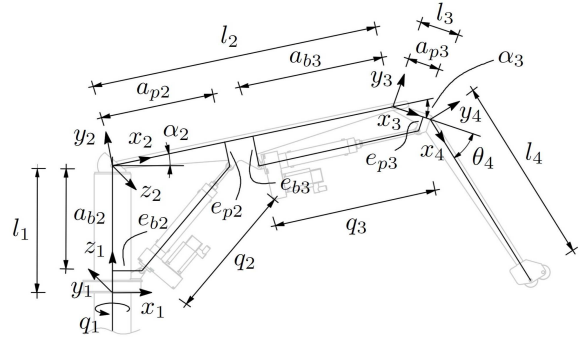}
\caption{Geometry of the knuckle boom crane.}
\label{kinematics_fig}
\end{figure}
In this work we consider a small-scale knuckle boom crane with a camera-based sensor package as shown in Fig.~\ref{fig:crane_system}. The crane has three actuated degrees of freedom (DOFs). The first actuated DOF is the slewing joint (1), which enables the crane king (4) to rotate about the vertical axis. The second actuated DOF is the extension of the linear actuator (2), which actuates the luffing of the inner boom (5). The third actuated DOF is the extension of the linear actuator (3), which is luffing the outer boom (6). The crane payload (9) is suspended from the crane tip by the cable (7). The payload used in the experiments is a hollow box with mass $m=\SI{12.7}{kg}$ and the dimensions are $223\times 223\times 241 \text{ mm}$. Two spherical markers (8) are attached to the cable (7). Three 2D cameras (10) are attached to the crane king (4), such that the cameras rotate with the slew motion of the crane. 
\subsection{Crane kinematics}\label{chap:CraneKinematics}
Consider the mechanical crane system given in Fig.~\ref{kinematics_fig} with the parameter values given in Table \ref{param}. We model the crane as an open-chain kinematic system, where the kinematic relations between the extensions of linear actuators and the joint rotations are defined. The crane pedestal is defined as Body $0$, the king is Body $1$, the inner boom is Body $2$ and the outer boom is Body $3$ and Body $4$. Each rigid Body $i$ has a body-fixed frame $i$. It is noted that frame $0$ is also the inertial frame. The length of Body $i$ is denoted $l_i$. The configuration of the system is defined by the vector of generalized coordinates
\begin{equation}
\begin{aligned}
\bm{q} = \begin{bmatrix}
\bm{q}_c^\mr{T} & \phi_x & \phi_y
\end{bmatrix}^\mr{T}, 
\label{q}
\end{aligned}
\end{equation}
where $\bm{q}_c = [q_1,q_2,q_3]^\mr{T}$.
The rotation matrix from frame $0$ to frame $1$ is $\bm{R}^0_1=\bm{R}_x(\pi) \bm{R}_z(-\frac{\pi}{2}) \bm{R}_z(q_1)$. The rotation matrix from frame $1$ to frame $2$ is $\bm{R}^1_2=\bm{R}_x(\frac{\pi}{2})\bm{R}_z(\alpha_2(q_2))$. The rotation matrix from frame $2$ to frame $3$ is $\bm{R}^2_3=\bm{R}_z(\alpha_3(q_3))$. The rotation matrix from frame $3$ to frame $4$ is $\bm{R}^3_4=\bm{R}_z(\theta_4)$. The origin of frame $5$ is attached to the crane tip, but the frame has the same orientation as the inertial frame, that is $\bm{R}^0_5=\bm{I}$. The rotation matrix from frame $5$ to frame $6$ is $\bm{R}^5_6=\bm{R}_x(\phi_x)\bm{R}_y(\phi_y)$, which also leads to $\bm{R}^0_6=\bm{R}_x(\phi_x)\bm{R}_y(\phi_y)$. Frame 6 is body-fixed frame of the payload cable.
The matrices $\bm{R}_x,\bm{R}_y,\bm{R}_z \in \text{SO(3)}$ are the orthogonal rotation matrices about the $x$, $y$ and $z$ axes, respectively \cite{Siciliano2008}. The angle $\alpha_i(q_i)$ is defined as 
\begin{equation}
\begin{aligned}
    \alpha_i(q_i) =& \arccos\frac{q_i^2 - b_{i1}^2 - b_{i2}^2}{-2 b_{i1} b_{i2}} 
    + \arctan \frac{e_{bi}}{a_{bi}} \\
    +& \arctan\frac{e_{pi}}{a_{pi}} - c_i, \ \ \text{for } i=2,3,
\label{alpha}
\end{aligned}
\end{equation}
where $c_2=0.5\pi$, $c_3=\pi$, $b_{i1}^2 = a_{bi}^2+e_{bi}^2$, $b_{i2}^2 = a_{pi}^2+e_{pi}^2$ and $a_{bi},a_{pi},e_{bi},e_{pi}$ are defined in Fig. \ref{kinematics_fig}. The rates of the orientation angles \eqref{alpha} can be found by time differentiation which gives $\dot{\alpha}_i= \dot{q}_i \ \partial \alpha_i/ \partial q_i$. The relative angular velocities between the frames are given as
\begin{equation}
\begin{aligned}
\bs{\omega}^1_{01} =& \ \dot{q}_1 \bm{z}^1_1, \ \ \ 
    \bs{\omega}^i_{i-1,i} = \ \dot{\alpha}_i \bm{z}^i_i, \ \  \text{for }i=2,3,
\end{aligned}
\label{omegas}
\end{equation}
where $\bm{z}^i_i = [0,0,1]^\mr{T}$ and $\bs{\omega}^4_{34} =  \bm{0}$. The distance vectors between the origins of the frames given in the coordinate of the local frame as
\begin{equation}
\begin{aligned}
    \bm{p}^0_{01} =& \ \bm{0}, \ \ \ 
    \bm{p}^1_{12} = \begin{bmatrix} 0 & 0 & l_1 \end{bmatrix}^\mr{T}, \\
    \bm{p}^i_{i,i+1} =& \begin{bmatrix} l_i & 0 & 0 \end{bmatrix}^\mr{T}, \ \ \text{for } i=2,3,4.
\label{pi,i,i+1}
\end{aligned}
\end{equation}
\begin{table}[!t]
\renewcommand{\arraystretch}{1.2}
\caption{Geometrical parameters of the crane}
\label{param}
\centering
\begin{tabular}{cc|cc|cc}
\hline\hline
\bfseries Term & \bfseries Value & \bfseries Term & \bfseries Value & \bfseries Term & \bfseries Value   \\
\hline
$l_1$ & \SI{0.711}{m} &     $e_{b2}$ & \SI{0.154}{m} &  $a_{p3}$ & \SI{0.167}{m}  \\
$l_2$ & \SI{1.500}{m} &     $a_{p2}$ & \SI{0.600}{m} &  $e_{p3}$ & \SI{0.076}{m}  \\
$l_3$ & \SI{0.205}{m} &    $e_{p2}$ & \SI{0.130}{m} &  $\theta_4$ & \SI{-39.4}{deg} \\
$l_4$ & \SI{0.992}{m} &    $a_{b3}$ & \SI{0.750}{m}    \\
$a_{b2}$ & \SI{0.550}{m} & $e_{b3}$ & \SI{0.160}{m}    \\
\hline\hline
\end{tabular}
\end{table}
The distance vector from the origin of frame $1$ to the origin of frame $5$ given in the coordinates of frame $0$ is found as
\begin{equation}
\begin{aligned}
    \bm{p}^0_{15} = \bm{R}^0_1 \big\{ \bm{p}^1_{12} + \bm{R}^1_2 \big[ \bm{p}^2_{23} + \bm{R}^2_3(\bm{p}^3_{34} + \bm{R}^3_4\bm{p}^4_{45}) \big] \big\},
\label{p015}
\end{aligned}
\end{equation}
the distance vector from the origin of frame $2$ to the origin of frame $5$ given in the coordinates of frame $0$ is found as
\begin{equation}
\begin{aligned}
    \bm{p}^0_{25} = \bm{R}^0_1 \bm{R}^1_2 \big[ \bm{p}^2_{23} + \bm{R}^2_3(\bm{p}^3_{34} + \bm{R}^3_4\bm{p}^4_{45}) \big]
\label{p025}
\end{aligned}
\end{equation}
and the distance vector from the origin of frame $3$ to the origin of frame $5$ given in the coordinates of frame $0$ is found as
\begin{equation}
\begin{aligned}
    \bm{p}^0_{35} = \bm{R}^0_1 \bm{R}^1_2  \bm{R}^2_3(\bm{p}^3_{34} + \bm{R}^3_4\bm{p}^4_{45}) .
\label{p035}
\end{aligned}
\end{equation}
The linear velocity of the origin of frame $5$ due to the rotations of the joints is given in the coordinates of the inertial frame as
\begin{equation}
\begin{aligned}
    \bm{v}^0_{05} = \begin{bmatrix} \bm{\hat{z}}^0_{1}\bm{p}^0_{15} &  \bm{\hat{z}}^0_{2}\bm{p}^0_{25} &
    \bm{\hat{z}}^0_{3}\bm{p}^0_{35}
    \end{bmatrix}
    \begin{bmatrix} \dot{q}_1 \\  \dot{{\alpha}}_2 \\ \dot{{\alpha}}_3
    \end{bmatrix},
\label{v115}
\end{aligned}
\end{equation}
where $\bm{\hat{\cdot}}$ denotes the skew-symmetric form of a vector, $\bm{{z}}^0_{1}=\bm{R}^0_1\bm{{z}}^1_{1}$, $\bm{{z}}^0_{2}=\bm{R}^0_2\bm{{z}}^2_{2}$ and $\bm{{z}}^0_{3}=\bm{R}^0_3\bm{{z}}^3_{3}$. The expression of the velocity \eqref{v115} can be written as
\begin{equation}
\begin{aligned}
    \bm{v}^0_{05} = \bm{J} \bm{\dot{q}}_c,
\label{v115_2}
\end{aligned}
\end{equation}
where the Jacobian
\begin{equation}
\begin{aligned}
    \bm{J} =  \begin{bmatrix} \bm{\hat{z}}^0_{1}\bm{p}^0_{15} &  \bm{\hat{z}}^0_{2}\bm{p}^0_{25} &
    \bm{\hat{z}}^0_{3}\bm{p}^0_{35}
    \end{bmatrix}
    \begin{bmatrix} 1 & 0 & 0 \\
   0 & \frac{\partial \alpha_2}{\partial q_2} & 0 \\
   0 &0 &  \frac{\partial \alpha_3}{\partial q_3}
    \end{bmatrix}
\label{Jxz}
\end{aligned}
\end{equation}
maps the rates of the generalized coordinates of the crane to the linear velocity of the crane tip.
\subsection{Payload modeling}
The crane payload is modeled as a spherical pendulum, where the mass of the cable is neglected and the mass $m$ of the payload is lumped at the end of the cable of length $L$. The payload has a body-fixed frame $6$, which has the origin coinciding with the origin of frame $5$. The equations of motion are derived using Kane's method \cite{kane1985dynamics}. Assume that velocity of the origin of frame $5$ relative to frame $0$ is given in the coordinates of frame $0$ as $\bm{\bar{v}}^0_{05} =   [{\dot{x}_5} , \dot{y}_5 ,  {\dot{z}_5}]^\mr{T}$, where we assume that ${\dot{z}_5}=0$. Then the velocity of the payload relative to frame $0$ and expressed in the coordinates of frame $0$ is 
\begin{align} 
\bm{v}^0_{p} = \bm{\bar{v}}^0_{05} + \bs{\hat{\omega}}^0_{06}  \bm{R}^0_6 \bm{p}^6_{L},
\label{v1p}
\end{align}
where $\bm{p}^6_{L} = [0,0,L]^\mr{T}$, $\bs{{\omega}}^0_{06}=\bs{{\omega}}^0_{56}$ is the angular velocity of frame $6$ relative to frame $0$ expressed in the coordinates of frame $0$, and $\bm{\hat{\cdot}}$ denotes the skew-symmetric form of a vector. The acceleration of the payload can be derived from \eqref{v1p} as $\bm{a}^0_{p} = d \bm{v}^0_{p} / dt$.
Provided that the partial velocities with respect to the pendulum generalized speeds are given as $\bm{v}^0_{p,1} = \partial \bm{v}^0_{p} / \partial \dot{\phi}_x$ and $\bm{v}^0_{p,2} = \partial \bm{v}^0_{p} / \partial \dot{\phi}_y$, then the equations of motion of the spherical pendulum are formulated as
\begin{equation}
\begin{aligned} 
 \bm{v}^0_{p,i} (- \bm{a}^0_{p}m + \bm{g}^0) = 0, \ \ \ \text{for } i=1,2,
\label{eqmo1}
\end{aligned}
\end{equation}
where $\bm{g}^0 = [0,0,mg]^\mr{T}$ is the force of gravity given in the coordinates of frame $0$ and $g$ is the acceleration of gravity.
The equations of motion \eqref{eqmo1} can explicitly be written as
\begin{equation}
\begin{aligned}
     \ddot\phi_x c_y + \omega_0^2 s_x &= 2\dot\phi_x\dot\phi_y s_y + \dot{v}_y c_x/L, \\
     \ddot\phi_y + \omega_0^2 c_xs_y &= - \dot\phi_x^2 s_y c_y - (\dot{v}_xc_y + \dot{v}_y s_x s_y)/L ,
     \label{eqmo2}
\end{aligned}
\end{equation}
where $\omega_0^2 = g/L$, $\dot{v}_x = \ddot x_5$, $\dot{v}_y = \ddot y_5$, $s_{i}=\sin \phi_{i}$ and $c_i=\cos \phi_i$.

\section{Estimation of payload motion}
\label{vision}
\subsection{Vision}
\begin{figure}
\centering
\includegraphics[width=0.47\textwidth]{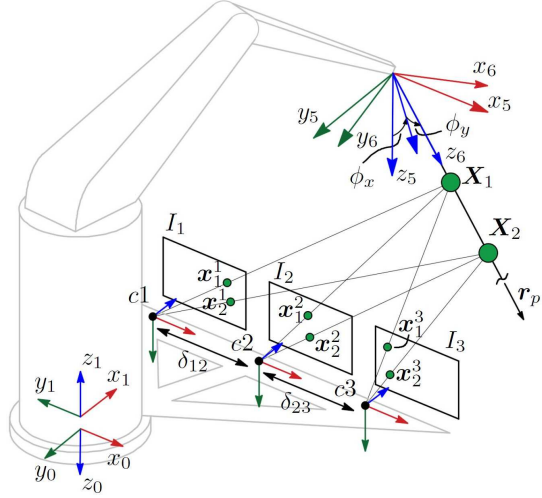}
  \caption{Three cameras are installed on a rack, which is rigidly attached to the king of the crane. The spherical markers are projected as pixels into the image planes of the cameras.}
  \label{cameras_fig}
\end{figure}
In this section we present the procedure for determination of the payload orientation angles $(\phi_x,\phi_y)$ (see Fig. \ref{cameras_fig}) using three-camera measurements. The measured angles are further used in combination with an extended Kalman filter for estimation of the orientation angles and angular velocities. Each camera $i$ has a camera-fixed frame $ci$.

Consider two points $\bm{X}_1=[X_1,Y_1,Z_1]^\tr$ and $\bm{X}_2=[X_2,Y_2,Z_2]^\tr$ given in Fig.~\ref{cameras_fig}, where each of the points is located in the center of a spherical marker. These points can be given relative to frame $c1$ of camera 1, and expressed in the coordinates of the inertial frame as
\begin{align}
    \bm{X}_j&=\bm{p}_{c1,5}^0 + \bm{R}^0_6 \Delta_j \bm{e}_3, \quad  \text{for }j=1,2,
\end{align}
where $\bm{p}_{c1,5}^0=\bm{p}_{05}^0-\bm{p}_{0,c1}^0$ is the vector from the origin of frame $c1$ to the origin of frame $5$, and $\bm{R}^0_6 \Delta_j \bm{e}_3$ is the position of the marker relative to the crane tip. The term $\Delta_j$ is the scalar distance from the origin of frame $5$ to the center of a spherical marker and $\bm{e}_3=[ 0 , 0 , 1 ]^\tr$.
Consider a line through the points $\bm{X}_j$ with a normalized direction vector $\bm{r}_p=[r_x,r_y,r_z]^\tr$ given as
\begin{align}
    \bm{r}_p&=\dfrac{\bm{X}_2-\bm{X}_1}{\Delta_2-\Delta_1}=\bm{R}^0_6 \bm{e}_3= \bb s_y & -s_xc_y & c_xc_y \eb^\tr. \label{eq:dirvec}
\end{align}
The points $\bm{X}_j$ are seen in the image plane $I_i$ of camera $i$ as pixels $\bm{x}^i_j=[u^i_j,v^i_j]^\tr$. The coordinates of the pixels in the image plane can be obtained as
\begin{align}
    \bm{\tilde{x}}^i_{j}=\bm{P}_i\bm{\tilde{X}}_j,\label{eq:meas}
\end{align}
where the tilde notation $\tilde{\cdot}$ is a homogeneous representation of a vector \cite{Siciliano2008}. 
The method for tracking and extracting the points $\bm{x}^i_j$ can be summed up by the following steps
\begin{itemize}
    \item Blur image $I_i$ by a Gaussian function;
    \item Convert $I_i$ from the RBG to the HSV color space and define a binary image $\hat{I}_i$ based on the range of pixel values in $I_i$;
    \item Apply the morphology operators erosion and dilation to $\hat{I}_i$;
    \item Enclose objects by circles and remove objects that are outside the range of desired radii;
    \item Find raw image moments of the objects and calculate the centroids $\bm{x}^i_1=[u^i_1,v^i_1]^\tr$ and $\bm{x}^i_2=[u^i_2,v^i_2]^\tr$ where $v^i_2>v^i_1$.
\end{itemize}
The camera matrix $\bm{P}_{i}$ for camera $i$ in (\ref{eq:meas}) is defined as
\begin{align}
    \bm{P}_{i}&=\bm{K}_{i}\left[\bm{R}^{ci}_{0}\, |\,  \bm{t}^{ci}_{ci,c1} \right],
 \end{align}
where the rotation matrix from the inertial frame to frame $ci$ is given as $\bm{R}^{0}_{ci}=\bm{R}^0_1\bm{R}_x(\pi/2)^\tr\bm{R}_y(\pi/2)$ for $i=1,2,3$ and $\bm{K}_i$ is the camera calibration matrix for camera $i$ \cite{Siciliano2008}.
The constant translation vectors from frame $ci$ to frame $c1$, expressed in the coordinates of $ci$ are $\bm{t}^{c2}_{c2,c1}=
-\delta_{12}\bm{e}_1$ and $\bm{t}^{c3}_{c3,c1}=-(\delta_{12}+\delta_{23})\bm{e}_1$, where $\bm{e}_1=[ 1 , 0 , 0 ]^\tr$.
The terms in (\ref{eq:meas}) will satisfy the equality $\bm{\hat{\tilde{x}}}^i_j \bm{P}_i\tilde{\bm{X}}_j = \bm{0}$.
Define $\bm{\bar{A}}^i_j = \bm{\hat{\tilde{x}}}^i_j \bm{P}_i \in \mathbb{R}^{3 \times 4}$, and let $\bm{{A}}^i_j \in \mathbb{R}^{2 \times 4}$ denote the matrix which is formed by removing the last row of $\bm{\bar{A}}^i_j$. Then $\bm{A}^i_j\tilde{\bm{X}}_j=\bm{0}$, where
\begin{align}
    \bm{A}^i_j=\begin{bmatrix}  v^i_j\bm{P}_i^3-\bm{P}_i^2  \\
                                \bm{P}_i^1 - u^i_j\bm{P}_i^3  \end{bmatrix},
                                \label{Aij}
\end{align}
and $\bm{P}^{k}_i$ denotes the $k$-th row of $\bm{P}_i$.
Similarly as in \eqref{Aij} the equality $\bm{A}_j\tilde{\bm{X}}_j= \bm{0}$ can be formulated taking into account measurement from all the cameras, where the matrix $\bm{A}_j$ is given as
\begin{align}
    \bm{A}_j=\begin{bmatrix}(\bm{A}^1_j)^\tr & (\bm{A}^2_j)^\tr & (\bm{A}^3_j)^\tr \end{bmatrix}^\tr.
\end{align}
The constraint equation $\bm{A}_j\tilde{\bm{X}}_j=\bm{0}$ requires that the matrices $\bm{A}_j\in \mathbb{R}^{6\times 4}$ have rank 3 if the points $\bm{x}^i_j$ are exact without noise. Singular value decomposition of $\bm{A}_j$ gives
\begin{align}
    \bm{A}_j=\sum^{4}_{k=1}\sigma^j_k\bm{u}^j_k(\bm{\nu}^j_k)^{\mathrm T},
\end{align}
where $\sigma^j_4=0$ if the points $\bm{x}^i_j$ are exact without noise and $\sigma^j_4>0$ if the points are noisy.
The measurements $\bm{\bar{X}}_j$ of the actual points $\bm{X}_j$ are given by
\begin{align}
    \bm{\tilde{\bar{X}}}_j=\lambda^j\bm{\nu}^j_4,
\end{align}
where $\lambda^j \neq 0$ are scaling factors and $\bm{\nu}^j_4=[X_j,Y_j,Z_j,1]^\tr / \lambda^j$. The measurement $\bm{\bar{r}}_p=[\bar{r}_x,\bar{r}_y,\bar{r}_z]^\tr$ of the direction vector $\bm{r}_p$ is given by
\begin{align}
    \bm{\bar{r}}_p= \dfrac{\bm{\bar{X}}_2-\bm{\bar{X}}_1}
    {\|\bm{\bar{X}}_2-\bm{\bar{X}}_1\|_2}.
\end{align}
The measurement $\bm{y}=[y_1,y_2]^\mr{T}$ of the pendulum orientation angles $(\phi_x,\phi_y)$ is then
\begin{align}
    y_1&\!=\!\arctan \bigg[-\frac{\bar{r}_y}{\bar{r}_z}\bigg], \ \  
    y_2\!=\!\arctan \bigg[ \frac{\bar{r}_x}{(\bar{r}_y^2+\bar{r}_z^2)^{1/2} } \bigg] \label{eq:measure:phix} ,
\end{align}
where (\ref{eq:dirvec}) is used. This solution is based on the direct linear transformation algorithm in \cite{Hartley2004}. 
\subsection{Extended Kalman Filter}
Consider the state vector given as
\begin{align}
    \bm{z}=\begin{bmatrix}  \phi_x &\phi_y &\dot{\phi}_x & \dot{\phi}_y & n_x & n_y \end{bmatrix}^{\mathrm T},
\end{align}
where $\phi_i$ and $\dot{\phi}_i$ are the pendulum orientation angles and angular velocities, while $n_i$ are the bias states due to calibration error.
The input vector is given as
\begin{align}
     \bm{a}=\begin{bmatrix} \dot{v}_x & \dot{v}_y \end{bmatrix}^{\mathrm T},
\end{align}
where $\dot{v}_x$ and $\dot{v}_y$ are the crane tip accelerations in $x$ and $y$ directions with respect to the inertial frame.
The spherical pendulum dynamics are assumed to be imposed by the white process noise $\bm{w}$ such that it can be written as a nonlinear stochastic system
\begin{align}
    \dot{\bm{z}}=\bm{f}(\bm{z},\bm{a},L)+\bm{w} ,
    \label{eq:linmod}
\end{align}
where $\bm{f}(\bm{z},\bm{a},L)$ is given by (\ref{eqmo2}) and the bias error $\bm{n}=[ n_x , n_y]^\mr{T}$ is modeled as random walk.
The measurement (\ref{eq:measure:phix}) is obtained at discrete times $t_k,t_{k+1},t_{k+2},...$ with a constant time step $\Delta t$ as $\bm{y}_k=\bm{y}(t_k)$. Using this measurement and discretization of (\ref{eq:linmod}) yields to
\begin{equation}
\begin{aligned}
    \bm{z}_{k+1}&=\bm{f}_k(\bm{z}_{k},\bm{a}_k,L)+\bm{w}_k , \\ 
    \bm{y}_k&=\bm{h}(\bm{z}_k)+\bm{v}_k,
     \label{eq:discrete}
\end{aligned}
\end{equation}
where
\begin{align}
    \bm{f}_k(\bm{z}_k,\bm{a}_k,L)=\bm{z}_k+\bm{f}(\bm{z}_k,\bm{a}_k,L)\Delta t
\end{align}
and
\begin{align}
    \bm{h}(\bm{z})= \bb \phi_x + n_x &  \phi_y + n_y  \eb^\tr.
\end{align}
The process and measurement noise are $\bm{w}_k\sim \mathcal{N}(0,\,\bm{Q})$ and $\bm{v}_k\sim \mathcal{N}(0,\,\bm{R})$ with covariance matrices $\bm{Q}$ and $\bm{R}$.
The extended Kalman filter \cite{Brown2012} is summarized in Algorithm \ref{array-sum} with transition and observation matrices
\begin{align}
    \bm{F}_k=\dfrac{\partial \bm{f}_k }{\partial \bm{z}}\Big|_{ \hat{\bm{z}}_{k},\bm{a}_{k}}, \quad \bm{H}=\bm{H}_k=\dfrac{\partial \bm{h} }{\partial \bm{z}}\Big|_{ \bar{\bm{z}}_{k}}.
\end{align}
The pendulum oscillation angles and angular velocities $[\phi_x, \phi_y, \dot\phi_x, \dot\phi_y]^\tr$ are then extracted from $\hat{\bm{z}}_k$.
\begin{algorithm}
\caption{Extended Kalman Filter Implementation}
\label{array-sum}
\begin{algorithmic}[1]
    \State $k=1$, $\hat{\bm{z}}_{k-1}=\hat{\bm{z}}_0$, $\hat{\bm{P}}_{k-1}=\hat{\bm{P}}_0$
    \Loop 
        \State $\bar{\bm{z}}_{k}=\bm{f}_{k-1}(\hat{\bm{z}}_{k-1},\bm{a}_{k-1},L)$
        \State $\bar{\bm{P}}_{k}=\bm{F}_{k-1}\hat{\bm{P}}_{k-1}\bm{F}_{k-1}^\tr+\bm{Q}$
        \State $\bm{K}_{k}=\bar{\bm{P}}_{k}\bm{H}^\tr\left(\bm{R}+\bm{H}\bar{\bm{P}}_{k}\bm{H}^\tr\right)^\inv$
        \State $\hat{\bm{z}}_{k}=\bar{\bm{z}}_{k}+\bm{K}_{k}\left(\bm{y}_{k}-\bm{H}\bar{\bm{z}}_{k}\right)$
        \State $\hat{\bm{P}}_{k}=\left(\bm{I}-\bm{K}_{k}\bm{H}\right)\bar{\bm{P}}_{k}$
        \State $k=k+1$
	\EndLoop
\end{algorithmic}
\label{alg:euler}
\end{algorithm}

\section{Cable length estimation}
\label{cable_length}
In this work the payload cable length $L$ is assumed to be the distance from the payload suspension point to the center of gravity of the payload. 

\begin{figure}
\centering
\includegraphics[width=0.47\textwidth]{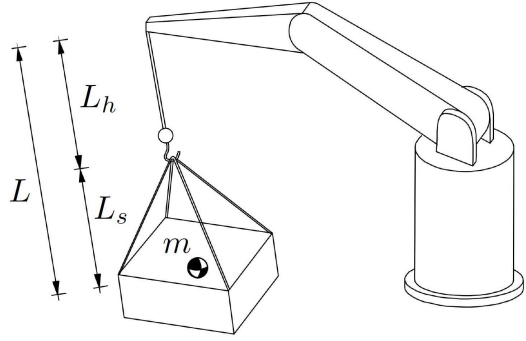}
\caption{The total cable length is a sum of crane cable outlet $L_h$ and the effective length of slings $L_s$.}
\label{cable_fig}
\end{figure}

Modern cranes are equipped with encoders for measuring the length of the released cable $L_h$ down to the hook, however, in practice, the payload is normally suspended from the hook using additional slings or chains, with unknown length $L_s$, as shown in Fig.~\ref{cable_fig}. In some cases $L_s$ can be significant in relation to $L_h$, which means that the total length of the payload cable $L$ in the pendulum model should be found as a sum of both. In this section we present the procedure for estimation of the total cable length $L$, which is required both in the control law and in the Kalman filter algorithm. We assume that the cable length $L$ is bounded by $L_{\min} \leq L \leq L_{\max}$. In fact, the dynamics of the pendulum in one plane is sufficient to estimate the length of the cable, therefore we propose to use only $\phi_x$ pendulum oscillations. 

Provided that $\phi_x$ is sufficiently small, we can linearize the first equation of \eqref{eqmo2} about the equilibrium point $(\phi_x, \dot\phi_x) = (0,0)$, which leads to
\begin{align}
    \ddot{\phi}_x = \dfrac{1}{L^*}(-g\phi_x+\dot v_y) \label{eq:adap},
\end{align}
where $L^*$ is the true unknown cable length. In our application its not possible to measure $\ddot{\phi}_x$ and the use of differentiation is not desirable. One way to solve it is to filter both side of (\ref{eq:adap}) with a $1$-order stable filter $1/\Lambda(s)$, where $\Lambda(s)=s+\lambda_0$ is a Hurwitz polynomial in $s$. Then the Laplace transformation of (\ref{eq:adap}) yields to the linear parametric model
\begin{align}
    z=\eta^* \psi \label{eq:z},
\end{align}
where $\eta^* =1/L^*$ and
\begin{equation}
\begin{aligned}
    z&=  [\dot{\phi}_x(s)s] / \Lambda(s), \\
     \psi &=[-g\phi_x(s)+\dot v_y(s)] / \Lambda(s).
\end{aligned}
\end{equation}
The variables $z$ and $\psi$ can be obtained without using differentiation. 
Consider $\eta (t)$ to be the estimate of $\eta^*$ at time $t$, then the estimated value $\hat{z}$ of the output $z$ is obtained as $\hat{z}=\eta\psi$.
Since the model (\ref{eq:z}) is an approximation of the true model (\ref{eqmo2}), we choose a least-square method for estimating $\eta$.
We introduce the normalized estimation error 
\begin{align}
    \epsilon=(z-\hat{z})/m_s^2=(z-\eta\psi)/m_s^2,
\end{align}
where $m_s^2=1+n_s^2$, the normalizing signal $n_s$ is chosen to be $n_s^2=\gamma\psi^2$ and $\gamma$ is a time-varying adaptive gain to be decided.
The optimal $\eta$ should minimize a cost function $J(\eta)$, where $\eta \in \mathcal{S} $ and $\mathcal{S}$ is a convex set given by
\begin{align}
    \mathcal{S}=\{\eta \in \mathbb{R} \,|\, g(\eta)\leq 0 \},
    \label{setS}
\end{align}
where $g: \mathbb{R} \to \mathbb{R}$ is a smooth function. In \cite{ioannou2012} the author suggested the following cost function
\begin{align}
    J(\eta)&=\frac{1}{2}\int^{t}_0e^{-\beta(t-\tau)}\epsilon^2(t,\tau)m_s^2(\tau)d\tau+\frac{1}{2\gamma_0}e^{-\beta t}(\eta-\eta_0)^2,
\end{align}
where $\epsilon(t,\tau)=\left(z(\tau)-\eta(t)\psi(\tau)\right)/m_s^2(\tau)$, the initial value of $\gamma$ is $\gamma_0>0$, the forgetting factor $\beta> 0$ and the initial value of $\eta$ is $\eta_0$. 
If estimates $\eta(t)$ are bounded by $1/L_{\max} \leq \eta \leq 1/L_{\min}$, then the new variable $\breve{\eta}=\eta-\eta_a$ is bounded by
\begin{align}
- \bar{\eta} \leq \breve{\eta} \leq \bar{\eta},
\label{breveeta}
\end{align}
where $\eta_a = 1/L_{\max} + \bar{\eta}$ and 
\begin{align}
\bar{\eta} = \frac{1}{2} \frac{L_{\max}-L_{\min}}{L_{\max} L_{\min}}.
\end{align}
The inequality \eqref{breveeta} can be re-written as $|\breve{\eta}| \leq \bar{\eta}$, which leads to that the inequality $\breve{\eta}^2 -\bar{\eta}^2  \leq 0$ is also satisfied. The inequality $\breve{\eta}^2 -\bar{\eta}^2  \leq 0$ can be explicitly written as
\begin{align}
\eta^2 - 2\eta \eta_a + \eta_a^2 - \bar{\eta}^2 \leq 0.
\label{breveeta2}
\end{align}
Provided that $g(\eta)$ should be not greater than zero \eqref{setS}, then we suggest that the left-hand side of \eqref{breveeta2} is a reasonable admissible function for $g(\eta)$, that is $g(\eta)=\eta^2 - 2\eta \eta_a + \eta_a^2 - \bar{\eta}^2$, which can alternatively be written as
\begin{align}
g(\eta)&=\eta^2-\eta \dfrac{L_{\max}+L_{\min}}{L_{\max}L_{\min}}+\dfrac{1}{L_{\max}L_{\min}},
\label{gradGeta}
\end{align}
then the gradient of \eqref{gradGeta} is defined as
\begin{align}
   \nabla g(\eta)=2\eta-\dfrac{L_{\max}+L_{\min}}{L_{\max}L_{\min}}.
\end{align}
In \cite{ioannou2012} the solution to the defined optimisation problem was called the least-squares algorithm with projection and was given by
\begin{align}
    \dot{\eta}=\begin{cases} \gamma\epsilon\psi  \quad &\text{if }g(\eta)<0  \\
    &\text{or if }g(\eta)=0 \text{ and } (\gamma\epsilon\psi)\nabla g(\eta)\leq 0\\
    0  \quad &\text{otherwise}
\end{cases}
\end{align}
and
\begin{align}
    \dot{\gamma}=\begin{cases} \beta \gamma -\gamma^2 \psi ^2/m_s^2 \quad &\text{if }g(\eta)<0  \\
    &\text{or if }g(\eta)=0 \text{ and } (\gamma\epsilon\psi)\nabla g(\eta)\leq 0\\
    0  \quad &\text{otherwise}.
\end{cases}
\end{align}
The initial guess of the pendulum length $L_0=1/\eta_0 $ should satisfy $L_{\min}\leq 1/\eta_0 \leq L_{\max}$. The performance of the cable length estimation algorithm is studied by the experiment. The estimate of the cable length $L=1/\eta$ is used in the control problem and the extended Kalman filter.

\section{Control}
 \label{control}
  The controller was designed with a payload damping controller in an inner loop, and a controller for the crane tip motion in an outer loop. We propose a controller 
 \begin{equation}
\begin{aligned}
    \ddot{x}_5 &= 2L\zeta\omega_0 \dot{\phi}_y+u_x, \\
    \ddot{y}_5&=- 2L\zeta\omega_0 \dot{\phi}_x+u_y
    \label{eq:feedback}
\end{aligned}
\end{equation}
with feedback from the angular rates $\dot{\phi}_x$ and $\dot{\phi}_y$, where the acceleration $(\ddot{x}_5,\ddot{y}_5)$ of the crane tip in the horizontal plane is the control variable, and $u_x$ and $u_y$ are the control variables of the outer control loop. Implementation issues related to the use of acceleration for the control variables are discussed at the end of this section. The closed loop dynamics of the payload linearized about $(\phi_x,\phi_y,\dot{\phi}_x,\dot{\phi}_y)=\bm{0}$ are found from (\ref{eqmo2}) and (\ref{eq:feedback}) to be 
 \begin{equation}
\begin{aligned}
    \ddot{\phi}_x+2\zeta\omega_0\dot{\phi}_x+\omega_0^2\phi_x&= \frac{u_y}{L}, \\
    \ddot{\phi}_y+2\zeta\omega_0\dot{\phi}_y+\omega_0^2\phi_y&= -\frac{u_x}{L}, 
    \label{eq:linear}
\end{aligned}
\end{equation}
It is seen that in a special case when $(u_x,u_y)=(0,0)$ the linearized closed loop system is two harmonic oscillators with undamped natural frequency $\omega_0$ and relative damping $\zeta$. In a general case when $(u_x,u_y)$ are not necessarily zero, the Laplace transform of the closed loop dynamics \eqref{eq:linear} gives
\begin{align}
    \phi_x(s)&=G(s)u_y(s), \quad  \phi_y(s)=-G(s)u_x(s),
    \label{phi_s}
\end{align}
where the transfer function $G(s)$ is
\begin{align}
    G(s)= \frac{1}{L(s^2+2\zeta \omega_0s+\omega_0^2)}.
\end{align}
Insertion  of \eqref{phi_s} into the Laplace transform of \eqref{eq:feedback} gives
\begin{align}
x_5(s)&= \frac{H(s)}{s^2}u_x(s), \quad y_5(s)=\frac{H(s)}{s^2}u_y(s), 
\label{x(s)}
\end{align}
where
\begin{align}
    H(s)&=1-2L\zeta\omega_0sG(s) 
    = \dfrac{s^2+\omega_0^2}{s^2+2\zeta\omega_0s+\omega_0^2}.
\end{align}
For frequencies $\omega \ll \omega_0$ it follows that $H(j\omega ) \to 1$ and \eqref{x(s)} simplifies to
\begin{align}
    x_5(s)&=\frac{1}{s^2}u_x(s), \quad y_5(s)=\frac{1}{s^2}u_y(s). 
\end{align}
The position of the crane tip can be controlled with a PD controller
\begin{equation}
\begin{aligned}
    u_x&=k_p(x_d-x_5)+k_d(\dot{x}_d-\dot{x}_5),  \\
    u_y&=k_p(y_d-y_5)+k_d(\dot{y}_d-\dot{y}_5) ,
    \label{eq:uxy}
\end{aligned}
\end{equation}
where $\bm{p}^0_{05}=[x_5,y_5,z_5]^\tr$ is the position of the crane tip relative to the inertial frame and $(x_d,y_d)$ is the desired position of the crane tip. The gains can be selected as $k_p=w_s^2$ and $k_d=2\zeta_s\omega_s$, where $\omega_s \ll \omega_0$ and $\zeta_s$ can be selected in the range $[0.7,1]$. The condition $\omega_s \ll \omega_0$ should be sufficiently well satisfied if $\omega_s = \omega_0 / k_s$, where $k_s \geq 5$ and $\omega_s$ is the bandwidth in the outer loop.

In practice it will not be possible to command the acceleration of the crane tip. The solution is to command the velocity instead, as the crane used in the experiments and most industrial cranes will have velocity control with the desired velocity is input variable. Therefore the acceleration input was converted to velocity inputs, as described in \cite{Rauscher2018}. This was done by integrating the two acceleration commands $\ddot x_5$ and $\ddot y_5$ in \eqref{eq:feedback} to velocity commands $w_x$ and $w_y$, and then using $w_x$ and $w_y$ as inputs to the velocity loops given by
\begin{align}
    \dot{w}_x=\ddot{x}_5&, \quad \dot{w}_y=\ddot{y}_5, \\
    \dot{v}_x=\frac{1}{T_v}(w_x-v_x)&, \quad \dot{v}_y= \frac{1}{T_v}(w_y-v_y) .
\end{align}
If the bandwidth $1/T_v$ of the velocity loop is sufficiently fast compared to the bandwidth of the damping controller, the resulting velocities $v_y$ and $v_y$ will be close to the velocity commands $w_x$ and $w_y$, and it follows that the accelerations $\ddot x_5$ and $\ddot y_5$ will be sufficiently close to the commanded accelerations $\dot{v}_x$ and $\dot{v}_y$.
The commanded velocities $v_x$ and $v_y$ are further transformed to the crane joint space according to
\begin{align}
     \dot{\bm{q}}_{\mathrm{com}}&= \bm{J}^\inv \bb v_x & v_y & 0 \eb^\tr,
\end{align}
where the Jacobian $\bm{J}$ is given in \eqref{Jxz} and $ \dot{\bm{q}}_{\mathrm{com}}$ is the commanded joint velocities of the crane.

\section{Experimental results}
\label{experiment}
\begin{figure}
\centering
\includegraphics[width=0.47\textwidth]{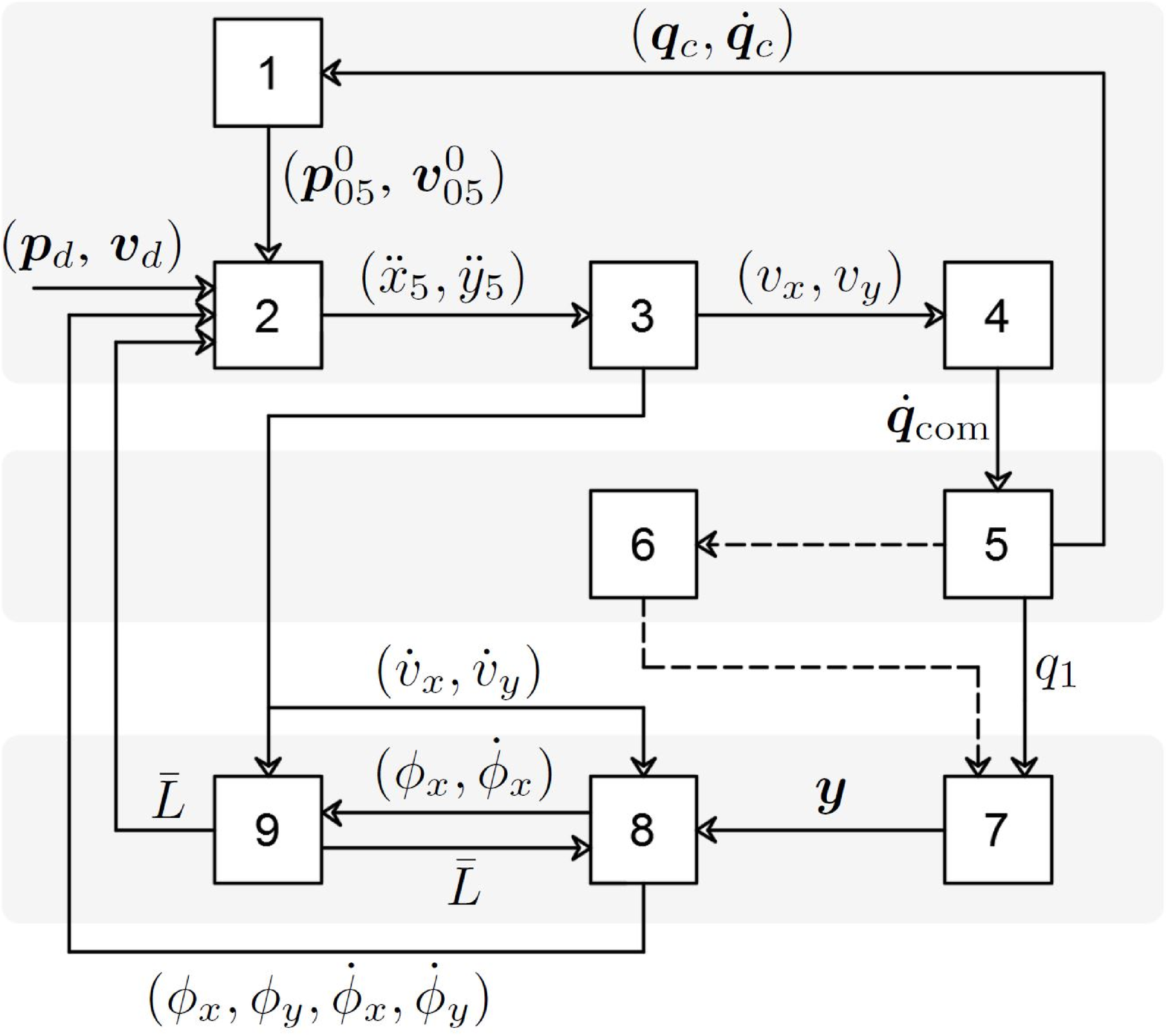}
\caption{Flow chart of the vision-based controller where (1) is the forward kinematics and Jacobian, (2) is the controller, (3) is the velocity loop, (4) is the inverse of the Jacobian, (5) is the physical crane, (6) is the physical payload, (7) is the vision system, (8) is the extended Kalman Filter and (9) is the cable length estimation.}
\label{signals_flow}
\end{figure}
The performance of the proposed mechatronic system was evaluated in laboratory experiments. A setup with a scaled knuckle boom crane was designed and constructed. The crane setup is shown in Fig.~\ref{fig:crane_system}. The crane was driven by one servo motor and two electro-mechanical cylinders (EMCs) driven by servo motors. All servo motors were equipped with encoders for measuring angles and angular velocities. The vision-based sensor system consisted of three consumer grade web cameras, where the resolution was selected as $1280\times720$ pixels. The distance between the cameras was $\delta_{12}=\delta_{23}= \SI{0.24}{m}$, and the spherical markers had a diameter of $\SI{0.03}{m}$. 
The control hardware consisted of a personal computer (PC) and a programmable logic controller (PLC). The PLC was used to read the data from the motor encoders and send commands to the servo drives to control the servo motor. The PC was used for computation and communication with the PLC. The measurements of $\bm{q}_c$ and  $\dot{\bm{q}}_c$ were obtained from the PLC. 
For the software part, MATLAB/Simulink was used on the PC for running the controller and the cable length estimation algorithm, while Python with OpenCV was used for the vision calculations and for the extended Kalman filter. The control period was set to 50 ms. The communication between MATLAB/Simulink and Python was implemented with UDP. The full overview of the signals in the system is given in Fig.~\ref{signals_flow}.

The covariance of the process noise in the extended Kalman filter was $\bm{Q}=10^{-4}\text{diag}(0.3,0.3,5,5,1,1)$ and the covariance of the measurement noise was 
\begin{align*} \quad \bm{R}=10^{-3}\bb 3.77597 &-2.10312 \\ -2.10312& 1.25147 \eb.
\end{align*}
 The initial a posteriori state was $\hat{\bm{z}}_0=\bm{0}$ and the error covariance matrix was $\hat{\bm{P}}_0=\bm{0}$.

 The forgetting factor in the cable length estimation algorithm was $\beta=0.5$ and the initial adaptive gain was $\gamma(0)=100$.
\subsection{Cable Length Estimation}
\label{experiment_cable}
The performance of the cable length estimation algorithm was studied in an experiment where the payload was oscillating and the crane tip position was stationary. The true cable length, which is the distance from the suspension point to the center of gravity of the payload, was $L^*=\SI{1.05}{m}$. In the experiments the payload was excited by manually applying initial displacements of different magnitude. During the experiment the amplitude of the oscillations was naturally damped by few degrees. The estimated cable length $L$ and the low-pass filtered estimate $\bar{L}$ were logged in the experiments, where the low-pass filtered estimate $\bar{L}$ of the cable length was used as an input to the extended Kalman filter. In all the tests the estimate of the cable length $L$ was bounded by $L_{\min}=\SI{0.3}{m}$ and $L_{\max}=\SI{1.5}{m}$. 

\begin{figure}[!t]
\centering
\includegraphics[width=0.47\textwidth]{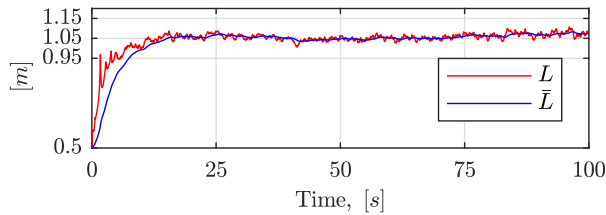}
\caption{Estimation of the cable length with $| \phi_x | < \SI{15}{deg}$, $L$ is the estimate and $\bar{L}$ is the estimate processed by a low-pass filter. The initial guess was $L_0 = \SI{0.5}{m}$, and the true length was $L^*=\SI{1.05}{m}$.}
\label{fig:ad_L05_15b}
\end{figure}

\begin{figure}[!t]
\centering
\includegraphics[width=0.47\textwidth]{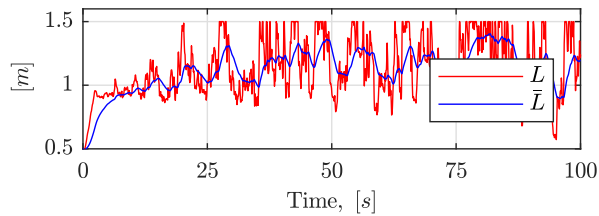}
\caption{Estimation of the cable length with $| \phi_x | < \SI{0.5}{deg}$ does not converge, $L$ is the estimate and $\bar{L}$ is the estimate processed by a low-pass filter.}
\label{fig:ad_L05_0}
\end{figure}

\begin{figure}[!t]
\centering
\includegraphics[width=0.47\textwidth]{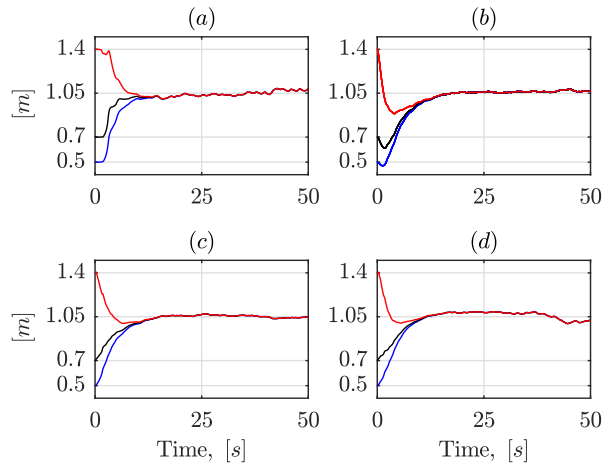}
\caption{Filtered estimates of the cable length with different initial guesses: (a) $| \phi_x | < \SI{5}{deg}$, (b) $| \phi_x | < \SI{10}{deg}$, (c) $| \phi_x | < \SI{15}{deg}$ and (d) $| \phi_x | < \SI{20}{deg}$.}
\label{fig:ad_L_58b15b20}
\end{figure}

In the first run, the initial amplitude of the payload oscillations was $\phi_x = \SI{15}{deg}$ and the initial cable length guess was $L_0=\SI{0.5}{m}$. The estimate of the cable length converged in less than \SI{10}{s}, as shown in Fig.~\ref{fig:ad_L05_15b}. Next, an experiment with the initial angle $\phi_x = \SI{0.5}{deg}$ was performed. In this case, the estimate of the length was more sensitive to noise, and the estimate oscillated between \SI{0.75}{m} and \SI{1.5}{m} after the initial convergence, as shown in Fig.~\ref{fig:ad_L05_0}. The reason for the loss of performance in this case is that the input data to the adaptive algorithm was not persistently exciting to achieve high quality in the estimates.  More test were run with initial angles of \SI{5}{deg}, \SI{10}{deg}, \SI{15}{deg} and \SI{20}{deg}, and initial cable length guesses of \SI{0.5}{m}, \SI{0.7}{m} and \SI{1.4}{m}, where the filtered cable length estimate $\bar{L}$ performed well with convergence after \SI{10}{s} and small variations after convergence (Fig.~\ref{fig:ad_L_58b15b20}).

The experimental results shown in Fig.~\ref{fig:ad_L05_15b}, \ref{fig:ad_L05_0} and \ref{fig:ad_L_58b15b20} demonstrate that the cable length estimate was converging to the true length $L^* = \SI{1.05}{m}$ after approximately \SI{10}{s} in all the presented cases, except for the case with close-to-zero payload oscillations. It is seen that for small payload oscillation angles the estimate was more noisy, as in Fig.~\ref{fig:ad_L_58b15b20}(a), which could happen due to that the Kalman filter estimates were more noisy for small angles. For large payload oscillations,  as in Fig.~\ref{fig:ad_L_58b15b20}(d), the estimate deviated more from the true value, which could happen due to that the linearized pendulum model was used for the cable length estimation algorithm. As expected, the cable length failed to converge when the payload did not oscillate. It is suggested that the cable length should be estimated $\textit{before}$ the controller starts damping the payload motion. It is feasible in practice, because in most of the cases the crane operator would do a manual maneuver before reaching the desired crane tip position over the landing site, then the payload motion can be damped right before the payload landing.
%
\subsection{Crane control}
The performance of the crane controller was investigated in an experiment that represented a realistic hoisting operation, where a payload is first carried over a landing site, and then the payload oscillations are damped out so that the payload can be landed safely. 
The experiment was executed in the following order. The crane tip was initially at the position $\bm{p}^0_{05}=[1.27,1.27] \text{ m}$, and the payload was manually excited. Then at $t=\SI{1}{s}$ the desired crane tip position was set to $\bm{p}_d=[0.70,1.80] \text{ m}$, and the crane moved to the desired position and finished the maneuver at $t=\SI{12}{s}$. The controller for payload damping was turned off during this maneuver. At time $t=\SI{20}{s}$ the payload damping controller was turned on. The cable length was estimated throughout the whole experiment until the damping controller was turned on at $t=\SI{20}{s}$, then the estimate was frozen, see the discussion in Section \ref{experiment_cable}. The initial estimate of the cable length was set to $L_0 = \SI{0.3}{m}$. The bandwidth in the outer control loop was selected as $\omega_s=\omega_0/5$ and the relative damping ratio was selected as $\zeta_s = 1$. Three different values $0.05$, $0.1$ and $0.2$ were used for the relative damping ratio $\zeta$ in the inner loop. %

\begin{figure}[!t]
\centering
\includegraphics[width=0.47\textwidth]{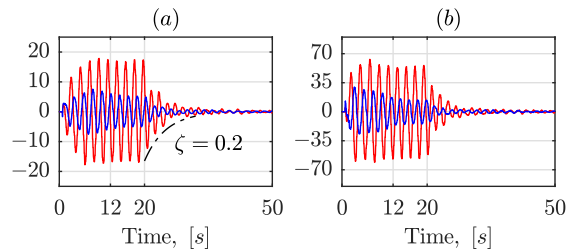}
\caption{Experimental results with $\zeta=0.2$: (a) orientation angles in [deg] and (b) angular velocities in [deg/s]. Red lines show $\phi_x$, $\dot{\phi}_x$, and blue lines show $\phi_y$, $\dot{\phi}_y$. The dashed line is a theoretical exponential decay curve.} 
\label{fig:phi_Z02_Fx18_Fy5_5g_lim03}
\end{figure}

\begin{figure}[!t]
\centering
\includegraphics[width=0.47\textwidth]{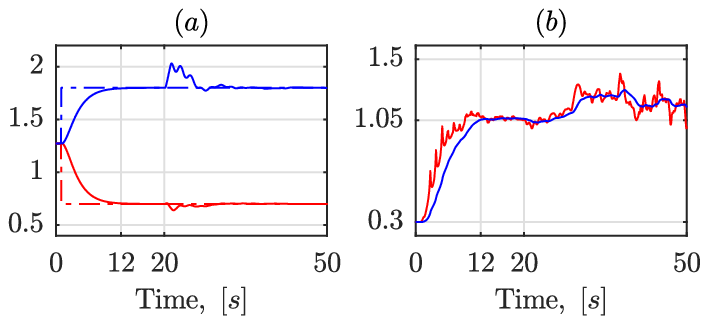}
\caption{Experimental results with $\zeta=0.2$: (a) crane tip position relative to the inertia frame in [m], red lines show $x$, $x_d$, and blue lines show $y$, $y_d$; (b) cable length estimate in [m], original estimate (in red) and filtered estimate (in blue).}
\label{fig:xL_Z02_Fx18_Fy5_5g_lim03}
\end{figure}

\begin{figure}[!t]
\centering
\includegraphics[width=0.47\textwidth]{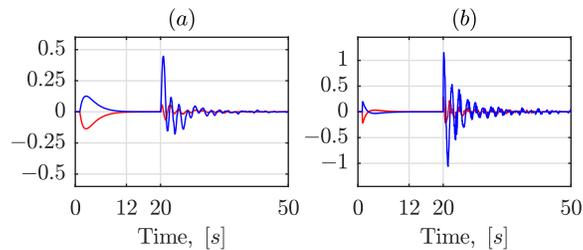}
\caption{Experimental results with $\zeta=0.2$: (a) commanded velocities of the crane tip in [m/s] and (b) control accelerations of the crane tip in [m/s$^2$]. Red lines show $v_x$, $\dot{v}_x$, and blue lines show $v_y$, $\dot{v}_y$.}
\label{fig:v_Z02_Fx18_Fy5_5g_lim03}
\end{figure}

\begin{figure}[!t]
\centering
\includegraphics[width=0.47\textwidth]{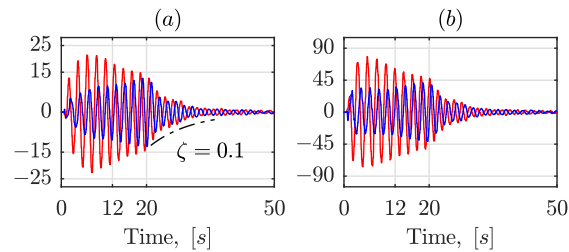}
\caption{Experimental results with $\zeta=0.1$: (a) orientation angles in [deg] and (b) angular velocities in [deg/s]. Red lines show $\phi_x$, $\dot{\phi}_x$, and blue lines show $\phi_y$, $\dot{\phi}_y$. The dashed line is a theoretical exponential decay curve.}
\label{fig:phi_Z01_Fx15_Fy13_5g_lim03}
\end{figure}

\begin{figure}[!t]
\centering
\includegraphics[width=0.47\textwidth]{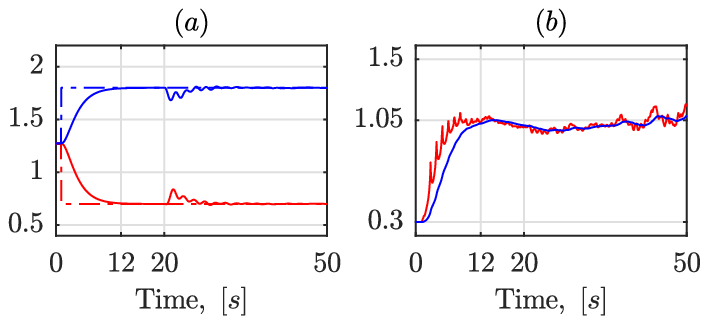}
\caption{Experimental results with $\zeta=0.1$: (a) crane tip position relative to the inertia frame in [m], red lines show $x$, $x_d$, and blue lines show $y$, $y_d$; (b) cable length estimate in [m], original estimate (in red) and filtered estimate (in blue).}
\label{fig:xL_Z01_Fx15_Fy13_5g_lim03}
\end{figure}

\begin{figure}[!t]
\centering
\includegraphics[width=0.47\textwidth]{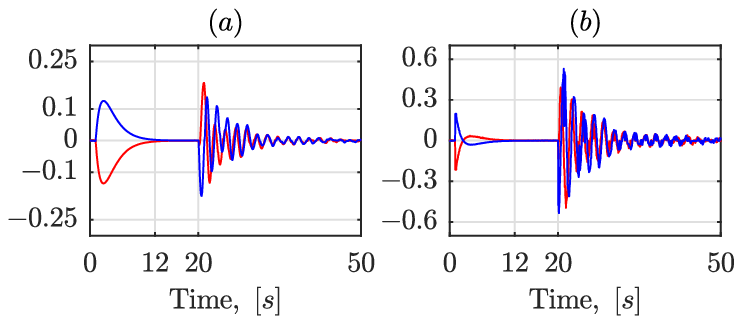}
\caption{Experimental results with $\zeta=0.1$: (a) commanded velocities of the crane tip in [m/s] and (b) control accelerations of the crane tip in [m/s$^2$]. Red lines show $v_x$, $\dot{v}_x$, and blue lines show $v_y$, $\dot{v}_y$.}
\label{fig:v_Z01_Fx15_Fy13_5g_lim03}
\end{figure}

\begin{figure}[!t]
\centering
\includegraphics[width=0.47\textwidth]{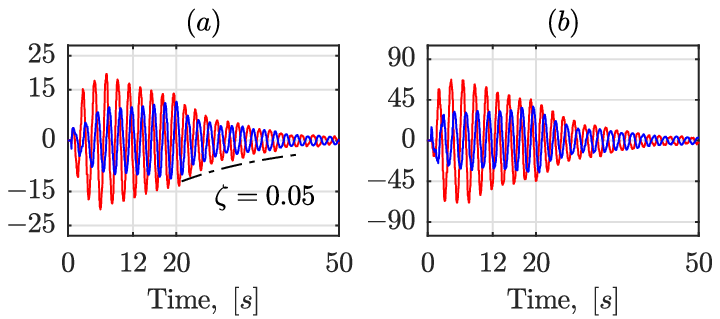}
\caption{Experimental results with $\zeta=0.05$: (a) orientation angles in [deg] and (b) angular velocities in [deg/s]. Red lines show $\phi_x$, $\dot{\phi}_x$, and blue lines show $\phi_y$, $\dot{\phi}_y$. The dashed line is a theoretical exponential decay curve.}
\label{fig:phi_Z005_Fx15_Fy10_5g_lim03}
\end{figure}

\begin{figure}[!t]
\centering
\includegraphics[width=0.47\textwidth]{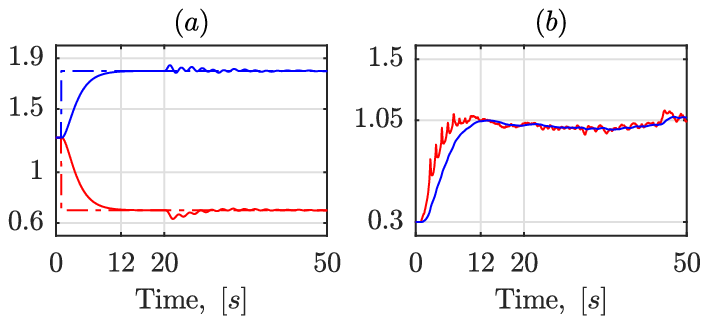}
\caption{Experimental results with $\zeta=0.05$: (a) crane tip position relative to the inertia frame in [m], red lines show $x$, $x_d$, and blue lines show $y$, $y_d$; (b) cable length estimate in [m], original estimate (in red) and filtered estimate (in blue).}
\label{fig:xL_Z005_Fx15_Fy10_5g_lim03}
\end{figure}

\begin{figure}[!t]
\centering
\includegraphics[width=0.47\textwidth]{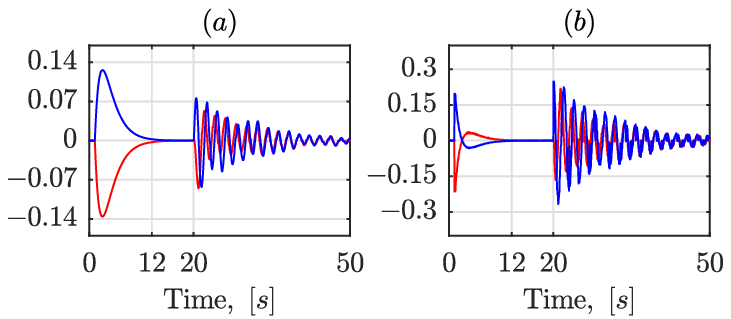}
\caption{Experimental results with $\zeta=0.05$: (a) commanded velocities of the crane tip in [m/s] and (b) control accelerations of the crane tip in [m/s$^2$]. Red lines show $v_x$, $\dot{v}_x$, and blue lines show $v_y$, $\dot{v}_y$.}
\label{fig:v_Z005_Fx15_Fy10_5g_lim03}
\end{figure}

The experimental results with the relative damping ratio $\zeta=0.2$ are given in Fig. \ref{fig:phi_Z02_Fx18_Fy5_5g_lim03}, \ref{fig:xL_Z02_Fx18_Fy5_5g_lim03} and \ref{fig:v_Z02_Fx18_Fy5_5g_lim03}. The values of the estimated payload orientation angles and angular velocities are shown in Fig. \ref{fig:phi_Z02_Fx18_Fy5_5g_lim03}. The values of the measured crane tip position relative to the inertial frame and given in the coordinates of the inertial frame, as well as the estimates of the cable length are given in Fig. \ref{fig:xL_Z02_Fx18_Fy5_5g_lim03}. The control acceleration and commanded velocities of the crane tip are given in Fig. \ref{fig:v_Z02_Fx18_Fy5_5g_lim03}.
The same types of experimental results with the relative damping ratio $\zeta=0.1$ are given in Fig. \ref{fig:phi_Z01_Fx15_Fy13_5g_lim03}, \ref{fig:xL_Z01_Fx15_Fy13_5g_lim03} and \ref{fig:v_Z01_Fx15_Fy13_5g_lim03} and the same types of experimental results with the relative damping ratio $\zeta=0.05$ are given in Fig. \ref{fig:phi_Z005_Fx15_Fy10_5g_lim03}, \ref{fig:xL_Z005_Fx15_Fy10_5g_lim03} and \ref{fig:v_Z005_Fx15_Fy10_5g_lim03}.

Given the results in Fig. \ref{fig:phi_Z02_Fx18_Fy5_5g_lim03}, \ref{fig:phi_Z01_Fx15_Fy13_5g_lim03} and \ref{fig:phi_Z005_Fx15_Fy10_5g_lim03}, the payload oscillations were controlled correctly with a decay of the oscillations close to theoretical curve for all tested $\zeta$ cases. In addition, the results in Fig. \ref{fig:xL_Z02_Fx18_Fy5_5g_lim03}, \ref{fig:xL_Z01_Fx15_Fy13_5g_lim03} and \ref{fig:xL_Z005_Fx15_Fy10_5g_lim03} show that the position of the crane tip was also controlled correctly, that is it eventually converged to the desired value. Both observations above let us conclude that the proposed cascade controller was efficient in all tested cases and the performance of the controller could be predicted by theoretical exponential decay curves. The performance of the proposed procedure for the cable length estimation can be evaluated from the results in Fig. \ref{fig:xL_Z02_Fx18_Fy5_5g_lim03}, \ref{fig:xL_Z01_Fx15_Fy13_5g_lim03} and \ref{fig:xL_Z005_Fx15_Fy10_5g_lim03}. It is noted that in contrast with the results in Fig. \ref{fig:ad_L_58b15b20}, here the oscillations were not free, that is the control acceleration was fed to the algorithm. In all tested cases the estimate of the cable length converged to the true value $L^*=\SI{1.05}{m}$ after 12 s and until the time $t=\SI{20}{s}$ the maximum error was $4.0$\%. As expected (see Section \ref{experiment_cable}), poorer convergence was demonstrated after the payload motion started being damped out at $t>\SI{20}{s}$.

\section{Conclusions}
\label{conclusions}
In this work we have presented a vision-based control system for a knuckle boom crane with online payload cable length estimation. The estimation of the payload oscillations was done using an extended Kalman filter with input from a visual sensor configuration that is novel for crane control. This visual sensor consisted of three 2D cameras rigidly attached to the crane king, which were tracking the position of two spherical markers on the payload cables. The markers were identified using the size and color information, where the color was selected to stand out from typical colors in the laboratory. No special background was used during the experiments and the markers were always correctly identified even when the background was unstructured laboratory equipment. The cable length estimation procedure was experimentally studied both for the case of free payload oscillations and for the case of forced oscillations. The convergence of the estimate was achieved with minor errors in all the cases, except for the case with close-to-zero oscillations, which was to be expected. The linear cascade controller designed for a linearized spherical pendulum model was experimentally verified using a realistic payload geometry and configuration. The experiments were conducted for three cases with a damping factor $\zeta=0.2$, $\zeta=0.1$ and $\zeta=0.05$. The controller efficiently damped out the payload oscillations and the crane tip position converged to the desired position in all the conducted experiments. The decay of the payload oscillations was very close to the theoretical exponential decay curves. 
\section*{Acknowledgment}
The research presented in this paper has received funding from the Norwegian Research Council, SFI Offshore Mechatronics, project number 237896.

\ifCLASSOPTIONcaptionsoff
  \newpage
\fi



%

\bibliographystyle{IEEEtran}
\bibliography{IEEEbiblio}

%
%

\end{document}